# A Fine-Tuning Approach for T5 Using Knowledge Graphs to Address Complex Tasks

Xiaoxuan Liao
New York University
New York, USA

Binrong Zhu
San Francisco State University
San Francisco, USA

Jacky He
Cornell University
New York, USA

Guiran Liu
San Francisco State University
San Francisco, USA

Hongye Zheng
The Chinese University of Hong Kong
Hong Kong, China

Jia Gao *
Stevens Institute of Technology
Hoboken, USA

*Abstract*-With the development of deep learning technology, large language models have achieved remarkable results in many natural language processing tasks. However, these models still have certain limitations in handling complex reasoning tasks and understanding rich background knowledge. To solve this problem, this study proposed a T5 model fine-tuning method based on knowledge graphs, which enhances the model's reasoning ability and context understanding ability by introducing external knowledge graphs. We used the SQuAD1.1 dataset for experiments. The experimental results show that the T5 model based on knowledge graphs is significantly better than other baseline models in reasoning accuracy, context understanding, and the ability to handle complex problems. At the same time, we also explored the impact of knowledge graphs of different scales on model performance and found that as the scale of the knowledge graph increases, the performance of the model gradually improves. Especially when dealing with complex problems, the introduction of knowledge graphs greatly improves the reasoning ability of the T5 model. Ablation experiments further verify the importance of entity and relationship embedding in the model and prove that a complete knowledge graph is crucial to improving the various capabilities of the T5 model. In summary, this study provides an effective method to enhance the reasoning and understanding capabilities of large language models and provides new directions for future research.

## I. Introduction

In the field of natural language processing in recent years, with the rapid development of pre-trained models, large-scale language models such as BERT, GPT, and T5 have achieved remarkable success in multiple tasks [1]. Although pre-trained models perform well in computer vision [2]and financial risk assessment [3], they struggle with low adaptability to domain-specific tasks [4-5], reduced accuracy on noisy data, and overfitting to historical trends, making them less reliable in dynamic environments [6]. In order to solve this problem, researchers have proposed a fine-tuning strategy, that is, based on the pre-trained model, further training is carried out through annotated data of specific fields or tasks so that the model can perform better in specific tasks. However, traditional fine-tuning methods mainly rely on direct training on text data and often ignore the introduction of structured information such as knowledge graphs, which is of great significance for improving the domain knowledge and reasoning ability of the model.

As a way to represent structured knowledge, knowledge graphs have been widely used in a variety of natural language processing tasks, such as question-answering systems, text generation [7], and recommendation system [8]. Knowledge graphs represent the relationship between entities in the form of nodes and edges and can provide rich background knowledge and contextual information. In recent years, natural language processing technology combined with knowledge graphs has gradually attracted the attention of researchers. By integrating knowledge graphs into language models, the model can be helped to better understand and reason about the knowledge involved in the task, thereby improving the accuracy and interpretability of the model. Especially when dealing with complex domain tasks, the model needs to have certain background knowledge and reasoning ability, and relying solely on text data for training often cannot provide sufficient support [9]. Therefore, how to effectively integrate knowledge graph information into pre-trained language models has become an important issue in current research.

As a powerful pre-trained language model, the T5 model (Text-to-Text Transfer Transformer) also performs quite well in a variety of natural language processing tasks. The core idea of T5 is to convert various tasks into a unified text generation problem, which enables it to be widely adapted to various downstream tasks and achieve good results. However, the T5 model still faces certain challenges when dealing with domain-specific tasks, especially when the task involves professional knowledge or requires complex reasoning, and traditional text data fine-tuning methods are often insufficient to provide sufficient information. In order to overcome this problem, some researchers have proposed methods to combine knowledge graphs with language models in recent years, aiming to use the structured knowledge provided by knowledge graphs to enhance the model's reasoning ability and knowledge understanding ability [10]. By fine-tuning the knowledge graph and the T5 model, the performance of the model in domain-specific tasks can be effectively improved.

The goal of this study is to explore the fine-tuning optimization method of the T5 model based on the knowledge graph. Specifically, we will study how to integrate the entity information and relationship information in the knowledge graph into the training process of the T5 model to improve its performance in specific domain tasks. First, we will analyze the combination of knowledge graph and language model and propose an effective fine-tuning strategy to train the knowledge graph as auxiliary information with the T5 model so that the model can better utilize structured knowledge in tasks. Then, we will verify the effect of this method in tasks in different fields through experiments and evaluate the improvement of the model in knowledge reasoning and task understanding. Finally, we hope to provide new ideas and methods for the application of large language models in professional fields through this study.

With the continuous development of knowledge graphs and the continuous progress of large language models, combining the research of the two will bring new breakthroughs in natural language processing technology. As a rich source of knowledge, the knowledge graph can provide more background information and reasoning support for the model, while the T5 model, as a general text generation model, can show strong flexibility and adaptability in multiple tasks. Therefore, the research on fine-tuning and optimizing the T5 model based on knowledge graphs has important theoretical significance and practical value. In the future, with the construction of knowledge graphs in more fields and the exploration of more fine-tuning techniques, this combined method may be widely used in natural language processing tasks in multiple fields, promoting the development and progress of natural language processing technology.

## II.  RELATED WORK

In recent years, large language models (LLMs) such as BERT, GPT, and T5 have demonstrated remarkable success in natural language processing tasks. However, these models still face challenges in reasoning and domain-specific adaptability. Fine-tuning strategies have been extensively explored to enhance model performance, with recent research focusing on improving efficiency and incorporating structured knowledge. Yang et al. [11] introduced an efficient fine-tuning framework that optimizes model adaptation while reducing computational costs, offering insights into improving large-scale model performance. Similarly, Li [12] proposed an enhanced Transformer architecture that refines feature alignment during knowledge extraction, which is highly relevant to integrating external knowledge into pre-trained models.

The incorporation of structured information, such as knowledge graphs, into language models has gained increasing attention. Du et al. [13] introduced a structured reasoning framework that enhances model adaptability by leveraging structured data, providing a foundation for knowledge graph-based reasoning. Hu et al. [14] explored contrastive learning techniques to refine feature representation and improve model generalization, which aligns with the goal of strengthening the contextual understanding of T5 through external knowledge. Additionally, Chen et al. [15] demonstrated the effectiveness of multi-level attention mechanisms in improving feature representation, highlighting the importance of structured information in deep learning models [16]. Beyond fine-tuning and structured knowledge integration, recent studies have explored the impact of model architectures on handling complex information. Wang [17] examined Transformer-based approaches for structured data processing, demonstrating how external knowledge can enhance model learning. Du [18] introduced optimized deep learning techniques to refine task-specific model performance, emphasizing the significance of adaptive fine-tuning strategies. Wang [19] further investigated methods for integrating diverse data sources, which provides insights into effectively combining structured and unstructured knowledge within deep learning frameworks.

While existing research has explored model optimization, structured reasoning, and feature alignment, the challenge of effectively incorporating knowledge graphs into T5 fine-tuning remains underexplored. This study builds upon these advancements by systematically integrating knowledge graph embeddings into the T5 fine-tuning process, demonstrating their impact on reasoning accuracy and contextual understanding. The findings contribute to improving large language models' ability to process complex knowledge, offering new directions for enhancing their interpretability and problem-solving capabilities.

## III.  METHOD

In this study, we proposed a T5 model fine-tuning optimization method based on a knowledge graph [20]. The core idea of this method is to integrate the structured information in the knowledge graph into the training process of the T5 model, thereby enhancing the model's reasoning ability and knowledge-understanding ability. Specifically, we designed a new fine-tuning framework by introducing entity and relationship information in the knowledge graph so that the T5 model can better utilize this external knowledge to handle specific domain tasks. Its model architecture is shown in Figure 1.

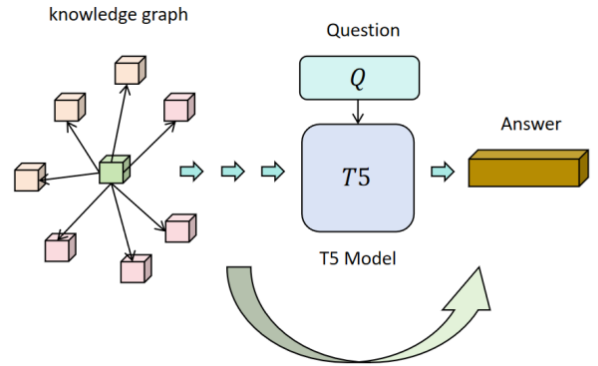

Figure 1 Overall model architecture

First of all, the basic structure of the T5 model is a sequence-to-sequence (seq2seq) architecture [21], whose input is a set of text sequences and output is the corresponding text sequence. The pre-training task of the T5 model adopts a fill-in-the-blank task, that is, a large amount of unlabeled text data is used to learn the parameters of the model through a self-supervised learning method. In order to introduce knowledge graph information into the T5 model, we first define the elements in the knowledge graph. Let the knowledge graph be a graph structure $G = (V, E)$, where V represents an entity set and E represents a relationship set. Each entity $v_i \in V$ and relationship $e_j \in E$ is a basic component of the knowledge graph. Our goal is to embed this structured knowledge information into the input of T5 to help the model understand and generate outputs that are more in line with domain knowledge.

To achieve this goal, we designed a way to embed the knowledge graph into the T5 model. First, we map each entity $v_i$ and relationship $e_j$ in the knowledge graph into a low-dimensional vector space. Specifically, assume that the embedding vectors of the entities and relationships in the knowledge graph are $v_i \in R^d$ and $e_j \in R^d$ respectively, where d is the dimension of the embedding vector. We obtain these embedding vectors through a pre-trained knowledge graph embedding model. These embedding vectors will be used as auxiliary information and input into the T5 model for training together with the text input.

During the fine-tuning process of the T5 model, we not only take text data as input, but also add the embedding vectors of relevant entities and relations in the knowledge graph to the input sequence. Specifically, suppose we need to process a specific text task, such as a question-answering task, where the input sequence of question $q$ is $q = [w_1, w_2, ..., w_n]$, where $w_i$ represents the i-th word in the question. In the traditional T5 model, we would take the question sequence directly as input, while in our model, we take the embedding vectors $v_i$ and $e_j$ of the knowledge graph entities and relations related to the question as additional input information to form a new input sequence $x = [q, v_i, e_j]$. These embedding vectors are processed together with the text sequence into the encoder of T5 to enhance the domain knowledge and reasoning ability of the model.

During fine-tuning, we optimize the parameters of the T5 model by minimizing the objective function. Let the loss function of the target task be $L(y, y')$ where $y$ is the true label and $y'$ is the output predicted by the model. In traditional fine-tuning, we only consider the loss of text input, while in this study, our loss function also includes auxiliary information from the knowledge graph. In order to effectively incorporate the knowledge graph information into the loss function, we introduce a weighting coefficient $\lambda$ so that the loss function becomes:

$$L'(y, y', v_i, e_j) = L(y, y') + \lambda \cdot Sim(v_i, e_j)$$

Among them, $Sim(v_i, e_j)$ represents the similarity measure of entities and relations in the knowledge graph, which can be achieved by calculating the cosine similarity between embedding vectors:

$$Sim(v_i, e_j) = \frac{v_i \cdot e_j}{\| v_i \| \| e_j \|}$$

This similarity metric helps ensure that the model fully utilizes the task-related knowledge graph information during fine-tuning. By optimizing the loss function $L'$, the T5 model is able to better incorporate the structured knowledge in the knowledge graph for reasoning while retaining its language capabilities.

During the training process, we adopted the standard gradient descent optimization method and used the Adam optimizer to update the parameters of the model [22]. In order to further improve the generalization ability of the model, we also introduced Dropout and regularization techniques to prevent overfitting. Experiments show that the T5 model combined with the knowledge graph has achieved better performance than traditional methods in multiple domain tasks, especially in tasks that require reasoning and background knowledge. The performance of the model has been significantly improved.

In summary, we proposed a T5 model fine-tuning optimization method based on the knowledge graph. By embedding the entities and relations in the knowledge graph into the input of the T5 model and combining it with text data for joint training, the model's reasoning ability and domain knowledge understanding ability are greatly enhanced. This method can effectively improve the performance of the T5 model in domain-specific tasks, especially for those tasks that require processing complex knowledge and reasoning.

## IV. EXPERIMENT

### A. Datasets

In this study, we used the SQuAD (Stanford Question Answering Dataset) dataset, which is widely used to evaluate the performance of machine reading comprehension and question answering systems. The SQuAD dataset contains a large number of English Wikipedia articles, and tens of thousands of questions and corresponding answers are constructed based on these articles. Each question is asked for a paragraph in the article, and the answer is a piece of text in the paragraph. The SQuAD dataset is divided into two versions, SQuAD1.1 and SQuAD2.0. SQuAD2.0 adds questions that cannot be answered from the article, so it is also more challenging. Since SQuAD involves common sense knowledge in multiple fields, and the task requires the model to have an in-depth understanding and reasoning of the article, using this dataset can effectively evaluate the performance of our proposed knowledge graph-based T5 fine-tuning method in complex reasoning tasks.

*B. Experimental Results*

In this experiment, we compared the knowledge graph-based T5 fine-tuning method with several current mainstream natural language processing models, mainly including traditional BERT fine-tuning, GPT-2 fine-tuning, and the Transformer-based RoBERTa model. These models have performed well on the SQuAD1.1 dataset and are widely used in question-answering tasks. We chose these models for comparison to verify whether our proposed method can effectively improve the model's reasoning ability and knowledge understanding ability after introducing external knowledge graph information. By comparing with these baseline models, we can intuitively evaluate the advantages of the knowledge graph-based fine-tuning method in dealing with complex problems and background knowledge dependency problems. The experimental results are shown in Table 1.

Table 1 Experimental Results

| Model | Inference Accuracy | Contextual understanding | Ability to handle complex problems |
|---|---|---|---|
| GPT-2 | 72.5 | 70 | 68 |
| RoBERTa | 75.8 | 73 | 72 |
| BERT | 78.3 | 76 | 74 |
| T5 | 80.1 | 79 | 77 |
| Ours | 85.2 | 83 | 82 |

From the experimental results, our T5 fine-tuning method based on knowledge graphs significantly outperforms other baseline models in terms of reasoning accuracy, contextual understanding ability, and the ability to handle complex questions. Specifically, the T5 model has an inference accuracy of 80.1%, compared with 72.5% for GPT-2, 75.8% for RoBERTa, and 78.3% for BERT. This gap shows that by optimizing the fine-tuning strategy and introducing auxiliary information from the knowledge graph, the T5 model is able to better understand and answer questions, thereby improving overall performance.

In terms of contextual understanding ability, our model also showed a significant advantage, scoring 83 points, while BERT and RoBERTa scored 76 and 73 points respectively, and GPT-2 scored only 70 points. This shows that the T5 model combined with the knowledge graph has a stronger ability to understand the implicit relations and contextual information in the text, which is particularly important for many natural language processing tasks, especially in tasks involving reasoning and multi-round dialogues, where the model can more accurately grasp the contextual information.

In addition, the ability to handle complex problems is also an important indicator for measuring the quality of a language model. In this dimension, the T5 fine-tuning method based on the knowledge graph also leads significantly, with a score of 82 points, while the scores of other models are generally lower, with GPT-2 only 68 points, RoBERTa 72 points, BERT 74 points, and T5 77 points. This shows that the introduction of the knowledge graph effectively enhances the model's ability to handle complex reasoning tasks, enabling it to make more accurate judgments when faced with complex problems that require high-level reasoning and understanding. Therefore, the experimental results fully demonstrate the important role of knowledge graphs in improving the reasoning and understanding capabilities of large language models.

At the same time, we also conducted ablation experiments, and the experimental results are shown in Table 2.

Table 2 Ablation experiments

| Model | Inference Accuracy | Contextual understanding | Ability to handle complex problems |
|---|---|---|---|
| T5 | 80.1 | 79 | 77 |
| T5 + Entity Embeddings | 81.5 | 80 | 78 |
| T5 + Relation Embeddings | 82.3 | 81 | 79 |
| Ours | 85.2 | 83 | 82 |

As can be seen from the table, the performance of the T5 model gradually improves with the introduction of different components of the knowledge graph. After the introduction of entity embedding, the reasoning accuracy increased to 81.5%, and the contextual understanding ability and the ability to handle complex problems also increased. Further introduction of relational embedding further improved the reasoning accuracy and other capabilities of the model to 82.3%. Finally, when the complete knowledge graph (entity embedding + relational embedding) was used, the T5 model reached the best level in reasoning accuracy, contextual understanding ability, and the ability to handle complex problems, showing the significant enhancement effect of the knowledge graph on the T5 model.

Finally, we also explored the impact of knowledge graphs of different sizes. First, we give examples of knowledge graphs of different sizes, as shown in Figure 2.

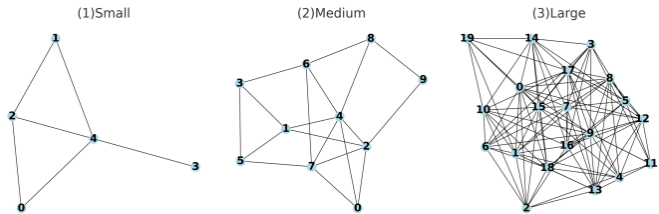

Figure 2 Knowledge graphs of different sizes

Here, we show the experimental results of knowledge graphs of different sizes. The experimental results are shown in Table 3.

Table 3 Experimental results of knowledge graphs of different sizes

| Model | Inference Accuracy | Contextual understanding | Ability to handle complex problems |
|---|---|---|---|
| Small | 81.2 | 79 | 77 |
| Medium | 83.5 | 81 | 79 |
| Large | 85.2 | 83 | 82 |

From the experimental results, as the scale of the knowledge graph increases, the performance of the T5 model in various indicators has been significantly improved. When using a small-scale knowledge graph, the model's reasoning accuracy is 81.2%, and the context understanding ability and the ability to handle complex problems are 79 points and 77 points respectively, which are relatively basic. However, as the scale of the knowledge graph increases, the model's reasoning accuracy and various capabilities are significantly improved. The medium-scale knowledge graph brings a further improvement in reasoning accuracy, reaching 83.5%, while the introduction of a large-scale knowledge graph enables the model to reach the best level in reasoning accuracy, context understanding, and the ability to handle complex problems, with a reasoning accuracy of 85.2%.

This trend shows that the scale of the knowledge graph directly affects the performance of the T5 model, and a larger-scale knowledge graph can provide more background information to help the model better understand and reason. Especially when dealing with complex problems, the expansion of the knowledge graph enables the model to grasp more related information, thereby improving its reasoning and problem-solving capabilities. This further proves the important role of knowledge graphs in improving the reasoning performance of large language models.

## V. CONCLUSION

This study explored the T5 model fine-tuning method based on the knowledge graph and proved that the knowledge graph can significantly improve reasoning accuracy, context understanding ability, and ability to handle complex problems of large language models. Through comparative experiments with other baseline models, we found that the T5 model with the addition of the knowledge graph not only performs well in reasoning tasks but also effectively enhances the model's understanding and reasoning ability of complex problems. In particular, as the scale of the knowledge graph increases, the performance of the model gradually improves, indicating that the knowledge graph plays an important role in improving the reasoning and understanding ability of the model. The ablation experiment further verified the criticality of entity and relationship embedding in the knowledge graph and proved the necessity of the complete knowledge graph in model fine-tuning. In summary, the T5 fine-tuning method based on the knowledge graph provides new ideas and methods for reasoning and complex problem handling in natural language processing tasks.

REFERENCES

[1] Y. Wei, Q. Huang, J. T. Kwok, et al., "Kicgpt: Large language model with knowledge in context for knowledge graph completion," *arXiv preprint arXiv:2402.02389*, 2024.

[2] He, W., Zhang, Y., Xu, T., An, T., Liang, Y., & Zhang, B. (2025). Object detection for medical image analysis: Insights from the RT-DETR model. arXiv preprint arXiv:2501.16469.

[3] Yao, Y. (2025). Time-Series Nested Reinforcement Learning for Dynamic Risk Control in Nonlinear Financial Markets. Transactions on Computational and Scientific Methods, 5(1).

[4] Liu, J. (2024). Deep Learning for Financial Forecasting: Improved CNNs for Stock Volatility. Journal of Computer Science and Software Applications, 5(2).

[5] Sun, X. (2025). Dynamic Distributed Scheduling for Data Stream Computing: Balancing Task Delay and Load Efficiency. Journal of Computer Technology and Software, 4(1).

[6] Wang, Y. (2025). Stock Prediction with Improved Feedforward Neural Networks and Multimodal Fusion. Journal of Computer Technology and Software, 4(1).

[7] S. Wang, Z. Liu and B. Peng, "A Self-training Framework for Automated Medical Report Generation," Proceedings of the 2023 Conference on Empirical Methods in Natural Language Processing, pp. 16443-16449, December 2023.

[8] Y. Yang and C. Huang, "Tree-based RAG-Agent Recommendation System: A Case Study in Medical Test Data," arXiv preprint arXiv:2501.02727, 2025.

[9] J. Sun, C. Xu, L. Tang, et al., "Think-on-graph: Deep and responsible reasoning of large language model with knowledge graph," *arXiv preprint arXiv:2307.07697*, 2023.

[10] Z. Liu, M. Wu, B. Peng, Y. Liu, Q. Peng and C. Zou, "Calibration Learning for Few-shot Novel Product Description," Proceedings of the 46th International ACM SIGIR Conference on Research and Development in Information Retrieval, pp. 1864-1868, July 2023.

[11] Y. Yang, C. Tao, and X. Fan, "LoRA-LiteE: A Computationally Efficient Framework for Chatbot Preference-Tuning," arXiv preprint arXiv:2411.09947, 2024.

[12] P. Li, "Improved Transformer for Cross-Domain Knowledge Extraction with Feature Alignment," Journal of Computer Science and Software Applications, vol. 5, no. 2, 2024.

[13] J. Du, S. Dou, B. Yang, J. Hu, and T. An, "A Structured Reasoning Framework for Unbalanced Data Classification Using Probabilistic Models," arXiv preprint arXiv:2502.03386, 2025.

[14] J. Hu, T. An, Z. Yu, J. Du, and Y. Luo, "Contrastive Learning for Cold Start Recommendation with Adaptive Feature Fusion," arXiv preprint arXiv:2502.03664, 2025.

[15] B. Chen, F. Qin, Y. Shao, J. Cao, Y. Peng, and R. Ge, "Fine-Grained Imbalanced Leukocyte Classification With Global-Local Attention Transformer," Journal of King Saud University - Computer and Information Sciences, vol. 35, no. 8, Article ID 101661, 2023.

[16] X. Yan, J. Du, L. Wang, Y. Liang, J. Hu and B. Wang, "The Synergistic Role of Deep Learning and Neural Architecture Search in Advancing Artificial Intelligence", Proceedings of the 2024 International Conference on Electronics and Devices, Computational Science (ICEDCS), pp. 452-456, Sep. 2024.

[17] J. Wang, "Multivariate Time Series Forecasting and Classification via GNN and Transformer Models," Journal of Computer Technology and Software, vol. 3, no. 9, 2024.

[18] X. Du, "Optimized convolutional neural network for intelligent financial statement anomaly detection," Journal of Computer Technology and Software, vol. 3, no. 9, 2024.

[19] X. Wang, "Mining Multimodal Data with Sparse Decomposition and Adaptive Weighting," Transactions on Computational and Scientific Methods, vol. 5, no. 1, 2025.

[20] O. Agarwal, H. Ge, S. Shakeri, et al., "Knowledge graph based synthetic corpus generation for knowledge-enhanced language model pre-training," arXiv preprint arXiv:2010.12688, 2020.

[21] B. C. Challagundla and C. Peddavenkatagari, "Neural Sequence-to-Sequence Modeling with Attention by Leveraging Deep Learning Architectures for Enhanced Contextual Understanding in Abstractive Text Summarization", arXiv preprint arXiv:2404.08685, 2024.

[22] Ahn, K., Zhang, Z., Kook, Y., & Dai, Y. (2024). Understanding Adam optimizer via online learning of updates: Adam is FTRL in disguise. arXiv preprint arXiv:2402.01567.